\documentclass[runningheads]{llncs}
\usepackage{graphicx}
\usepackage{comment}
\usepackage{amsmath,amssymb} 
\usepackage{color}
\usepackage{array,multirow,graphicx}
\usepackage{shortbold}
\usepackage{subcaption}
\usepackage{cite}
\usepackage{float}
\DeclareMathOperator*{\argmax}{arg\,max}
\usepackage[table,x11names]{xcolor}
\definecolor{highlightblue}{rgb}{0.8,0.93333,1.0}
\definecolor{highlightgreen}{rgb}{0.8,1.0,0.8}
\usepackage[width=122mm,left=12mm,paperwidth=146mm,height=193mm,top=12mm,paperheight=217mm]{geometry}

\newcommand*\yellowcircle[1]{\protect\includegraphics[width=3mm, trim=0.5mm 1.5mm 0.5mm 0.5mm]{figures/#1-yellow.pdf}}

\begin{document}
\pagestyle{headings}
\mainmatter
\def\ECCVSubNumber{5113}  

\title{DHOG: Deep Hierarchical Object Grouping} 

\titlerunning{DHOG: Deep Hierarchical Object Grouping}
\authorrunning{L. Darlow, A. Storkey}
\author{
Luke Nicholas Darlow\inst{1} \\
{\email{L.N.Darlow@sms.ed.ac.uk}}\\
\and
Amos Storkey\inst{1} \\
\email{a.storkey@ed.ac.uk}
}
{\institute{$^\text{1}$ School of Informatics\\
University of Edinburgh \\
10 Crichton St, Edinburgh EH8 9AB}}

\maketitle
\begin{abstract}
Recently, a number of competitive methods have tackled unsupervised representation learning by maximising the mutual information between the representations produced from augmentations. The resulting representations are then invariant to stochastic augmentation strategies, and can be used for downstream tasks such as clustering or classification. Yet data augmentations preserve many properties of an image and so there is potential for a suboptimal choice of representation that relies on matching easy-to-find features in the data. We demonstrate that greedy or local methods of maximising mutual information (such as stochastic gradient optimisation) discover local optima of the mutual information criterion; the resulting representations are also less-ideally suited to complex downstream tasks. Earlier work has not specifically identified or addressed this issue. We introduce deep hierarchical object grouping (DHOG) that computes a number of distinct discrete representations of images in a hierarchical order, eventually generating representations that better optimise the mutual information objective. We also find that these representations align better with the downstream task of grouping into underlying object classes. We tested DHOG on unsupervised clustering, which is a natural downstream test as the target representation is a discrete labelling of the data. We achieved new state-of-the-art results on the three main benchmarks without any prefiltering or Sobel-edge detection that proved necessary for many previous methods to work. We obtain accuracy improvements of: $4.3\%$ on CIFAR-10, $1.5\%$ on CIFAR-100-20, and $7.2\%$ on SVHN.
\end{abstract}

\section{Introduction}\label{sec:introduction}


It is very expensive to label a dataset with respect to a particular task. Consider the alternative where a user, instead of labelling a dataset, specifies a simple set of class-preserving transformations or `augmentations'. For example, lighting changes will not change a dog into a cat. Is it possible to learn a model that produces a useful representation by leveraging a set of such augmentations? Such a representation would need to be good at capturing salient information about the image, and enable downstream tasks to be done efficiently. If the representation were a discrete labelling which groups the dataset into clusters, an obvious choice of downstream task is unsupervised clustering that ideally should match the clusters that would be obtained by direct labelling, without ever having been learnt on explicitly labelled data. 



Using data augmentations to drive unsupervised representation learning has been explored by a number of authors \cite{dosovitskiy2014discriminative, dosovitskiy2015discriminative, bachman2019learning, chang2017deep, wu2019deep, ji2019invariant, cubuk2019randaugment}. These approaches typically involve learning neural networks that map augmentations of the same image to similar representations. This is a reasonable approach to take as the variances across many common image augmentations often align with the invariances we would require a method to have.

In particular, a number of earlier works target maximising mutual information (MI) between augmentations \cite{oord2018representation, hjelm2018learning, wu2019deep, ji2019invariant, bachman2019learning}. By targeting high MI between representations computed from distinct augmentations of images, useful representations can be learned that capture the invariances induced by the augmentations. We are particularly interested in a form of representation that is a discrete labelling of the data, as this is particularly parsimonious. This labelling can be seen as a clustering~\cite{ji2019invariant} procedure, where MI can be computed and assessment can be done directly using the learned labelling, as opposed to via an auxiliary network trained posthoc.

\subsection{Suboptimal mutual information maximisation}

We argue and show that in many cases the MI objective is not maximised effectively in existing work due to the combination of:
\begin{enumerate}
   
    \item \textbf{Greedy optimisation algorithms} used to train neural networks, such as stochastic gradient descent (SGD) that potentially target local optima; and
     \item A limited set of data augmentations that can result in the existence of multiple \textbf{local optima to the MI maximisation} objective.
\end{enumerate}

SGD is greedy in the sense that early-found high-gradient features can dominate and so networks will tend to learn easier-to-compute locally-optimal representations (for example, one that can be computed using fewer neural network layers) over those that depend on complex features. 

By way of example, in natural images, average colour is an easy-to-compute characteristic, whereas object type is not. If the augmentation strategy preserves average colour, then a reasonable mapping need only compute average colour information, and high MI between images representations will be obtained.

\subsection{Dealing with greedy solutions}

A number of heuristic solutions, such as as Sobel edge-detection~\cite{caron2018deep, ji2019invariant} as a pre-processing step, have been suggested to remove/alter the features in images that may cause trivial representations to be learned. However, this is a symptomatic treatment and not a solution. In the work presented herein, we acknowledge that greedy SGD can get stuck in local optima of the MI maximisation objective because of limited data augmentations. Instead of trying to prevent a greedy solution, we let our DHOG model learn this representation, but \emph{also require it to learn a second distinct representation}. Specifically, we minimise the MI between these two representations so that the latter cannot rely on the same features used by an earlier head. We extend this idea by adding additional representations, each time requiring the latest to be distinct from all previous representations.

\subsection{Contributions}

Learning a set of representations by encouraging them to have low MI, while still maximising the original augmentation-driven MI objective for each representation, is the core idea behind \textbf{D}eep \textbf{H}ierarchical \textbf{O}bject \textbf{G}rouping (DHOG). We define a mechanism to produce a set of hierarchically-ordered solutions (in the sense of easy-to-hard orderings, not tree structures). DHOG is able to better maximise the original MI objective between augmentations since each representation must correspond to a unique local optima. Our contributions are:



\begin{enumerate}
\item We identify the \textbf{suboptimal MI maximisation problem}: maximising MI between data augmentations using neural networks and stochastic gradient descent (SGD) produces a substantially suboptimal solution.\footnote{We show this by finding higher mutual information solutions using DHOG, rather than by any analysis of the solutions themselves.} We reason that the SGD learning process settles on easy-to-compute solutions early in learning and optimises for these as opposed to leveraging the capacity of deep and flexible networks. We give plausible explanations and demonstrations for why this is a case, and show, with improved performance on a clustering task, that we can explicitly avoid these suboptimal solutions. 
\item We mitigate for this problem, introducing DHOG: the first robust neural network image grouping method to learn diverse and hierarchically arranged \textit{sets of discrete image labellings} (Section \ref{sec:method}) by explicitly modelling, accounting for, and avoiding spurious local optima, requiring only simple data augmentations, and needing no Sobel edge detection.
\item We show a marked improvement over the current state-of-the-art for standard benchmarks in image clustering for CIFAR-10 ($4.3\%$ improvement), CIFAR-100-20 (a 20-way class grouping of CIFAR-100, $1.5\%$ improvement), and SVHN ($7.2\%$ improvement); we set a new accuracy benchmarks on CINIC-10; and show the utility of our method on STL-10 (Section \ref{sec:experiments}).
\end{enumerate}

To be clear, DHOG still learns to map data augmentations to similar representations as this is imperative to the learning process. The difference is that the DHOG framework enables a number of intentionally distinct data labellings to be learned, arranged hierarchically in terms of source feature complexity. 

\subsection{Assessment task: clustering}
For this work, our focus is on finding higher MI representations; we then assess the downstream capability on the ground truth task of image classification, meaning that we can either (1) learn a representation that must be `decoded' via an additional learning step, or (2) produce a discrete labelling that requires no additional learning. Clustering methods offer a direct comparison and require \emph{no labels for learning a mapping from the learned representation to class labels}. Instead, labels are only required to appropriately assign groups to appropriate classes and no learning is done using these. Therefore, our comparisons are with respect to state-of-the-art clustering methods.

\section{Related Work}\label{sec:related}
The idea of MI maximisation for representation learning is called the \textit{infoMAX} principle~\cite{linsker1988self, tschannen2019mutual}. Contrastive predictive coding~\cite{oord2018representation} (CPC) models a 2D latent space using an autoregressive model and defines a predictive setup to maximise MI between distinct spatial locations. Deep InfoMAX~\cite{hjelm2018learning} (DIM) does not maximise MI across a set of data augmentations, but instead uses mutual information neural estimation~\cite{belghazi2018mine} and negative sampling to balance maximising MI between global representations and local representations. Local MI maximisation encourages compression of spurious elements, such as noise, that are inconsistent across blocks of an image. Augmented multiscale Deep InfoMAX~\cite{bachman2019learning} (AMDIM) incorporates MI maximisation across data augmentations and multiscale comparisons. 

\subsection{Clustering}
Clustering approaches are more directly applicable for comparison with DHOG because they explicitly learn a discrete labelling. The authors of deep embedding for clustering (DEC)~\cite{xie2016unsupervised} focused their attention on jointly learning an embedding suited to clustering and a clustering itself. They argued that the notion of distance in the feature space is crucial to a clustering objective. Joint unsupervised learning of deep representations and image clusters (JULE)~\cite{yang2016joint} provided supervisory signal for representation learning. Some methods \cite{ghasedi2017deep,fard2018deep} employ autoencoder architectures along with careful regularisation of cluster assignments to (1) ensure sufficient information retention, and (2) avoid cluster degeneracy (i.e., mapping all images to the same class). 

Deep adaptive clustering~\cite{chang2017deep} (DAC) recasts the clustering problem as binary pairwise classification, pre-selecting comparison samples via feature cosine distances. A constraint on the DAC system allows for a one-hot encoding that avoids cluster degeneracy. Another mechanism for dealing with degeneracy is to use a standard clustering algorithm, such as $K$-means to iteratively group on learned features. This approach is used by DeepCluster~\cite{caron2018deep}. 

Associative deep clustering (ADC)~\cite{haeusser2018associative} uses the idea that associations in the embedding space are useful for learning. Clustering was facilitated by learning a network to associate data with (pseudo-labelled) centroids. They leveraged augmentations by encouraging samples to output similar cluster probabilities. 

Deep comprehensive correlation mining~\cite{wu2019deep} (DCCM) constructs a sample correlation graph for pseudo-labels and maximises the MI between augmentations, and the MI between local and global features for each augmentation. While many of the aforementioned methods estimate MI in some manner, invariant information clustering~\cite{ji2019invariant} (IIC) directly defines the MI using the $c$-way softmax output (i.e., probability of belong to class $c$), and maximises this over data augmentations to learn clusters. They effectively avoid degenerate solutions because MI maximisation implicitly targets marginal entropy. We use the same formulation for MI -- details can be found in Section \ref{sec:method}.

\section{Method}\label{sec:method}
\begin{figure}[!htb]
\begin{center}
\includegraphics[trim={ 0 0 10mm 0},clip, width=\linewidth]{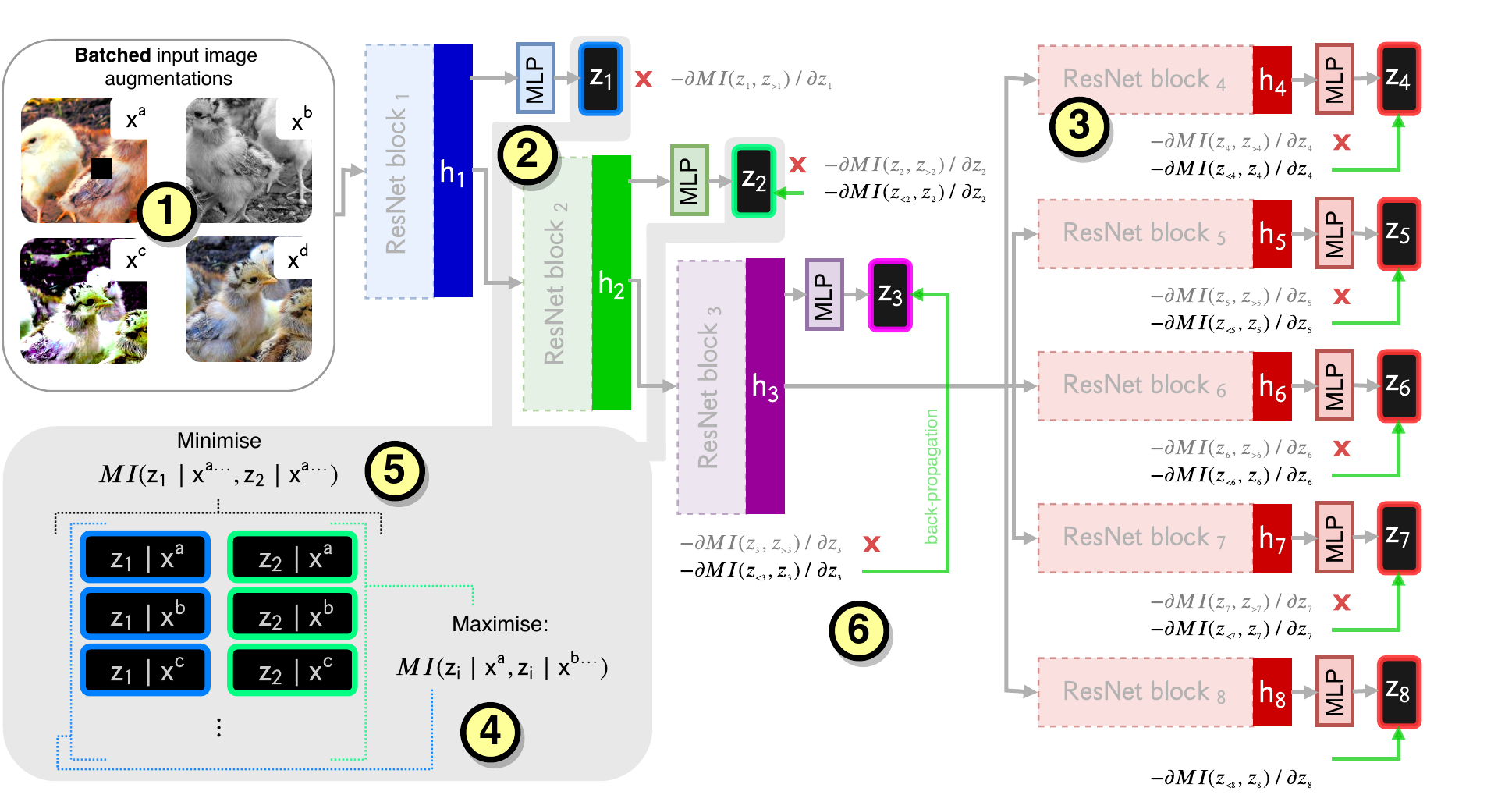}

  \caption{DHOG architecture. The skeleton is a ResNet18~\cite{he2016identity}. The final ResNet block is repeated $k-3$ times ($k=8$ here). \yellowcircle{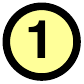}  Augmentations of each image, $\xRV^{a\ldots d}$, are separately processed by the network. \yellowcircle{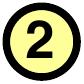} Each shallow ResNet block ($1\ldots3$) constitutes shared computation for deeper blocks, while also computing separate probability vectors, $\zRV_1 \ldots \zRV_3$. Each $\zRV_i$ is viewed as the probability for each outcome of the random variable $c_i$ that makes a discrete labelling choice. \yellowcircle{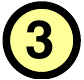} The deepest ResNet blocks compute further $\zRV_{>3}$. \yellowcircle{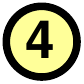} The network is trained by maximising the MI between allocations $c_i$ from \emph{all data augmentations}, and \yellowcircle{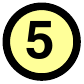} separately for each node $i$, minimising the MI between $c_i$ and $c_{< i}$ for the \emph{same data augmentation}. \yellowcircle{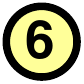} This is implemented as a global optimisation by stopping gradients such that they are \emph{not back-propagated} for later computation paths.
  \label{fig:architecture}}
 \end{center}

\end{figure}
DHOG is an approach for obtaining jointly trained multi-level representations as discrete labellings, arranged in a simple-to-complex hierarchy. Later representations in the hierarchy have low MI between earlier representations. 

Each discrete labelling is computed by a separate `head' within a neural network -- Figure \ref{fig:architecture} shows the architecture. A head is an unit that computes a multivariate class probability vector. By requiring independence amongst heads, a diversity of solutions to the MI maximisation objective can be found. The head that best maximises MI between augmentations typically aligns better with a ground truth task that also relies on complex features (e.g., classification).

Figure \ref{fig:architecture} demonstrates the DHOG architecture and training principles. There are shared model weights (\yellowcircle{2}: ResNet blocks 1, 2, and 3) and head-specific weights (the MLP layers and \yellowcircle{3}: ResNet blocks 4 to 8). For the sake of brevity, we abuse notation and use $\mathrm{MI}(\zRV,\zRV')$ between labelling probability vectors as an overloaded shorthand for the mutual information $\mathrm{MI}(c,c')$  between the labelling random variables $c$ and $c'$ that have probability vectors $\zRV$ and $\zRV'$ respectively.



Any branch of the DHOG architecture (\yellowcircle{1} to any $\zRV_i$) can be regarded as a single neural network. These are trained to maximise the MI between the label variables at each head for different augmentations; i.e. between label variables with probability vectors $\zRV_i(\xRV)$ and $\zRV_i(\xRV^{\prime})$ for augmentations $\xRV$ and $\xRV^{\prime}$. Four augmentations are shown at \yellowcircle{1}. The MI is maximised pairwise between all pairs, at \yellowcircle{4}. This process can be considered \emph{pulling} the mapped representations together.

Following IIC~\cite{ji2019invariant}, we compute the MI directly from the label probability vectors within a minibatch. Let $\zB_{i},\zB_{i}^\prime$ denote the random probability vectors at head $i$ associated with sampling a data item and its augmentations, and passing those through the network. 
Then we can compute the mutual information between labels associated with each augmentation using
\begin{equation}
    \begin{split}
        \mathrm{MI}_{aug}(c_i, c_i^\prime)& = \mathrm{Tr} (E[\zB_{i}(\zB_{i}^\prime)^T]^T \log (E[\zB_{i}(\zB_{i}^\prime)^T]))  \\ & \quad- E[\zB^T_{i}] \log E[\zB_{i}] - E[(\zB^\prime_{i})^T] \log E[(\zB^\prime_{i})]
    \label{eq:MI}
    \end{split}
\end{equation}{}
where $\mathrm{Tr}$ is the matrix trace, logarithms are computed element-wise, and expectations are over data samples and augmentations of each sample. In practice we compute an empirical estimate of this MI based on samples from a minibatch.




\subsection{Distinct heads}\label{sec:distinct}
What makes DHOG different from other methods is that each head is encouraged to compute unique solutions via \emph{cross-head MI minimisation}. For a minibatch of images, the particular labelling afforded by any head is trained to have low MI with other heads -- \yellowcircle{5} in Figure \ref{fig:architecture}. We assume multiple viable groupings/clusterings because of natural patterns in the data. By encouraging low MI between heads, these must capture different patterns in the data.

Concepts such as brightness, average colours, low-level patterns, etc., are axes of variation that are reasonable to group by, and which maximise the MI objective to some degree. Complex ideas, such as shape, typically require more processing. Greedy optimisation may not discover these groupings without explicit encouragement. Unfortunately, the groupings that tend to rely on complex features are most directly informative of likely class boundaries. In other words, without a mechanism to explore viable patterns in the data (like our cross-head MI minimisation, Section \ref{sec:mimin}), greedy optimisation will avoid finding them.

\subsection{Cross-head MI minimisation}\label{sec:mimin}
Our approach to addressing \emph{suboptimal MI maximisation} is to encourage unique solutions at sequential heads ($\zRV_1 \ldots \zRV_8$ in Figure \ref{fig:architecture}), which rely on different features in the data. We can compute and minimise the MI across heads using:
\begin{equation}
    \begin{split}
        \mathrm{MI}_{head}(c_i, c_j)& = \mathrm{Tr}(E[\zRV_i\zRV_j^T]^T \log (E[\zRV_i\zRV_j^T]))  \\ & \quad- E[\zRV_{i}^T] \log (E[\zRV_{i}]) - E[(\zRV_{j})^T] \log (E[\zRV_{j}]).
    \label{eq:MI2}
    \end{split}
\end{equation}{}
Logarithms are element-wise, and expectations are over the data and augmentations. Note $\zRV_i$ and $\zRV_j$ are each computed from the \emph{same data augmentation}. We estimate this from each minibatch sample. This process can be thought of as \emph{pushing} the heads apart.

\subsection{Aligning assignments}
When computing and subsequently minimising Equation \ref{eq:MI2}, a degenerate solution must be accounted for: two different heads can effectively minimise this form of MI computation while producing identical groupings of the data because of the way MI is computed in practice. This can be done simply by ensuring the order of the labels is permuted for each head, but consistently so across the data. We use the Hungarian Method~\cite{kuhn1955hungarian} to choose the best match between heads, effectively mitigating spurious MI computation. This step can be computationally expensive when $c$ is large, but is imperative to the success of DHOG.

\subsection{Hierarchical arrangement}\label{sec:arrange}
Requiring $k$ heads (where $k=8$ here) to produce unique representations is not necessarily the optimal method to account for suboptimal MI maximisation. Instead, what we do is encourage a simple-to-complex hierarchy structure to the heads, defined according to cross-head comparisons made using Equation \ref{eq:MI2}. The hierarchy enables a reference mechanism through which later representations \emph{can be encouraged toward relying on complex and abstract features in the data}. 

Figure \ref{fig:architecture} shows 8 heads, three of which are computed from representations owing to early residual blocks of the network. The hierarchical arrangement is created by only updating head-specific weights according to comparisons made with earlier heads. In practice this is done by stopping the appropriate gradients during training -- \yellowcircle{6} in the figure. For example, when computing the MI between assignments using $\zRV_{i=6}$ and those using $\zRV_{i\neq6}$, gradient back-propagation is allowed when $i < 6$ but not when $i > 6$. In other words, when learning to produce $\zRV_{i=6}$, the network is encouraged to produce a head that is distinct from heads `lower' on the hierarchy. Those `higher' on the hierarchy do not affect the optimisation, however. Extending this concept for $i=1\ldots8$ gives rise to the idea of the hierarchical arrangement.

Initial experiments showed clearly that if this hierarchical complexity routine was ignored, the gains owing to cross-head MI minimisation were reduced. 

\subsection{Objective}
The part of the objective producing high MI representations by `pulling' together discrete labellings from augmentations is Equation \ref{eq:MI} normalised over $k$ heads:
\begin{equation}
    \mathrm{MI}_{pull} = \frac{1}{k}\sum^k_{i=0}MI_{aug}(c_i, c_i^\prime).
    \label{eq:pull}
\end{equation}
The quantity used to `push' heads apart is Equation \ref{eq:MI2} normalised per head:
\begin{equation}
    \mathrm{MI}_{push} = \sum^k_{i=1}\frac{\sum^i_{\substack{j=1\\ j\neq i}}MI_{head}(c_i, c_j)}{i},
    \label{eq:push}
\end{equation}
where each cross-head MI term is scaled by the head index, $i$, since that directly tracks the number of comparisons made for each head. $i$ scales up the hierarchy, such that the total $\mathrm{MI}_{head}$ associated with any head is scaled according to \emph{the number of comparisons}. Scaling ensures that head-specific weight updates are all equally important. The final optimisation objective is:


\begin{equation}
    \theta^*=\argmax_{\theta} \mathrm{MI}_{pull} - \alpha \mathrm{MI}_{push},
    \label{eq:objective}
\end{equation}
where $\theta$ are the network parameters implicitly included in the MI computations, $\alpha$ is a hyper-parameter we call the \textit{cross-head MI-minimization coefficient}. We ran experiments in Section \ref{sec:experiments} with $\alpha=0$ as an ablation study.

\subsection{Design and training choices}\label{sec:archchoice}

Figure \ref{fig:architecture} shows the architecture and training design. It is based on a ResNet-18 backbone, where each residual block has two layers (with a skip connection over these). Blocks 1 to 3 have 64, 128, and 256 units, respectively. Each parallel final block (4 to 8, here) have 512 units. Each MLP has a single hidden layer of width 200. Although the parallel repetition of entire block structures for each head is cumbersome, our earlier experiments showed that this was an important model flexibility. We used four data augmentation repeats with a batch size of 220.

DHOG maximises MI between discrete labellings from different data augmentations. This is equivalent to a clustering and is similar to IIC. There are, however, key differences. In our experiments:
\begin{itemize}
    \item We train for \textbf{1000 epochs} with a cosine annealing learning rate schedule, as opposed to a fixed learning rate for 2000 epochs.
    \item We \textbf{do not use sobel edge-detection} or any other arbitrary preprocessing as a fixed processing step.
    \item We make use of the fast auto-augment CIFAR-10 data augmentation strategy (for all tested datasets) found by \cite{lim2019fast}. We then randomly apply (with $p=0.5$) grayscale after these augmentations, and take random square crops of sizes 64 and 20 pixels for STL-10 and all other datasets, respectively.
\end{itemize}{}

The choice of data augmentation is important, and we acknowledge that for a fair comparison to IIC the same augmentation strategy must be used. The ablation of any DHOG-specific loss (when $\alpha = 0$) largely recreates the IIC approach but with augmentations, network and head structure matched to DHOG; this enables a fair comparison between an IIC and DHOG approach. 

Since STL-10 has much more unlabelled data of a similar but broader distribution than the training data, the idea of `overclustering' was used by \cite{ji2019invariant}; they used more clusters than the number of classes (70 versus 10 in this case). We repeat each head with an overclustering head that does not play a part in the cross-head MI minimisation. The filter widths are doubled for STL-10. We interspersed the training data evenly and regularly through the minibatches.




To determine the DHOG cross-head MI-minimisation coefficient, $\alpha$, we carried out a non-exhaustive hyper parameter search using only CIFAR-10 images (without the labels), assessing performance on a held out validation set sampled from the training data. This did not use the evaluation data in any way.

\subsection{Assessment}

Once learned, the optimal head can be identified either using the highest MI, or using a small set of labelled data. Alternatively all heads can be used as different potential alternatives, with posthoc selection of the best head according to some downstream task. The union of all the head probability vectors could be used as a compressed data representation. In this paper the head that maximises the normalised mutual information on the training data is chosen. \emph{This is then fully unsupervised}, as with the head selection protocol of IIC. We also give results for the best posthoc head to show the potential for downstream analysis.\vspace{-1mm}












\vspace{5mm}
\section{Experiments}\label{sec:experiments}
We first show the functionality of DHOG using a toy problem (Section \ref{sec:toyexp}) to illustrate how DHOG can find distinct and viable groupings of the same data. In Section \ref{sec:related} we give results to show the superiority of DHOG on real images. \vspace{-1mm}

\subsection{Toy problem}\label{sec:toyexp}
\begin{figure}[!htbp]
\begin{center}
\begin{subfigure}{0.48\textwidth}
\includegraphics[width=0.98\linewidth]{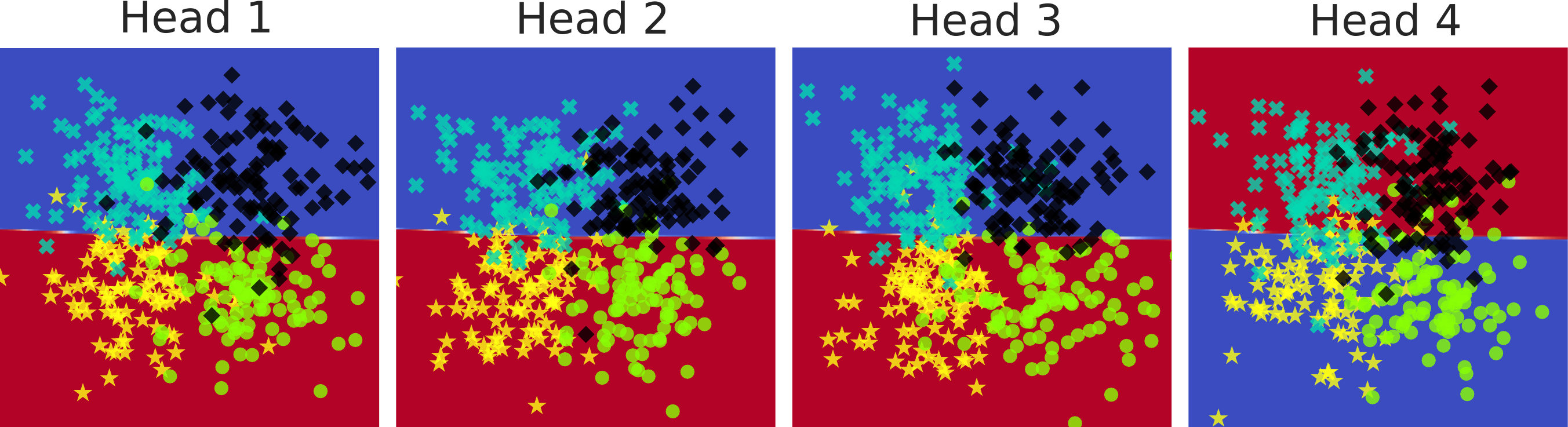}
\caption{Without DHOG training}
\end{subfigure}
\begin{subfigure}{0.48\textwidth}
\includegraphics[width=0.98\linewidth]{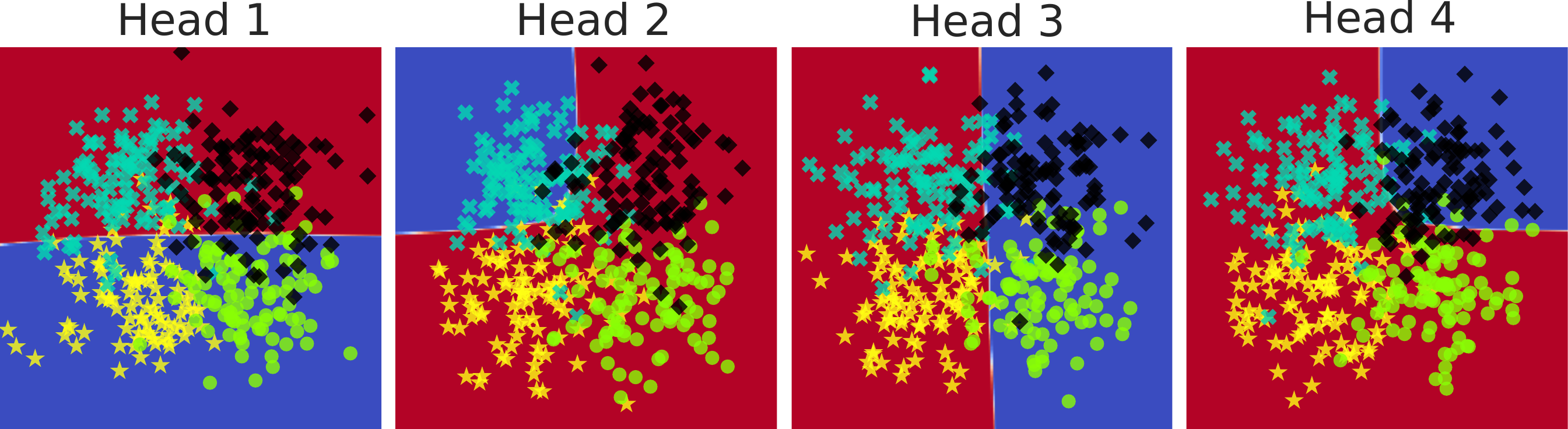}
\caption{With DHOG training}
\end{subfigure}
\end{center}
  \caption{Toy Problem: 4 sets of 2D Gaussian distributed points. The network must learn 2 groups. The probability of being in either is given by the background. Without DHOG the network simply learns a single solution, while DHOG encourages a number of unique solutions, from which it can select the solution with highest mutual information. }
\label{fig:toy}
\end{figure}

Figure \ref{fig:toy} demonstrates a simple 2D toy problem to show how DHOG can learn a variety of solutions. The data is generated from 4 separate 2D Gaussians: there are four possible groups. The number of clusters was set as 2 in order to create a circumstance where different groupings were possible. The augmentations were generated by adding additional Gaussian noise to the samples. Without DHOG (a) the network computes the same solution for each head. With DHOG (b) each head produces a unique solution. Even though the single solution found in (a) might very well be the sought after solution, it is not necessarily the solution that maximises the MI between augmentations of data. This is evidence of the \emph{suboptimal MI maximisation problem}: the network, learned using SGD, latched onto this local minima througout training. DHOG is able to identify different, better, minima because the local minima are already identified.

\subsection{Real images}\label{sec:real}

The datasets used for assessment were CIFAR-10~\cite{krizhevsky2009learning}, CIFAR-100-20 (CIFAR-100~\cite{krizhevsky2009learning} where classes are grouped into 20 super-classes), CINIC-10~\cite{darlow2018CINIC} (an extension of CIFAR-10 using images from Imagenet~\cite{imagenet_cvpr09} of similar classes), street view house numbers~\cite{netzer2011reading} (SVHN), and STL-10~\cite{coates2011analysis}. For CINIC-10 only the standard training set of 90000 images (without labels) was used for training.

\begin{table*}[!htbp]
\centering
\def\arraystretch{0.95}
\begin{tabular}{ccccc}
&\textbf{Method} & \textbf{Accuracy} & \textbf{NMI($\zRV$, $\yRV$)} & \textbf{ARI}\\ \hline
{{\begin{tabular}{c}\parbox[t]{2mm}{\multirow{12}{*}{\rotatebox[origin=c]{90}{\tiny\textbf{CIFAR-10}}}}\end{tabular}}}
& $K$-means on pixels & $ 21.18 \pm 0.0170$ & $0.0811 \pm 0.0001$ & $0.0412 \pm 0.0001$ \\
& Cartesian K-means~\cite{wang2014optimized} & $22.89$ & $0.0871$ & $0.0487$ \\
& JULE~\cite{yang2016joint} & $27.15$ & $0.1923$ & $0.1377$ \\
& DEC~\cite{xie2016unsupervised} $^\dagger$ & $30.1$ &-&-\\
& DAC~\cite{chang2017deep} & $52.18$ & $0.3956$ & $0.3059$ \\
& DeepCluster~\cite{caron2018deep}  $^\dagger$ & $37.40$ &-&-\\
& ADC~\cite{haeusser2018associative} & $29.3 \pm 1.5$ &-&-\\
& DCCM~\cite{wu2019deep} & $62.3$ & $0.496$  & $0.408$ \\
&IIC~\cite{ji2019invariant} $^\dagger$ & $57.6 \pm 5.01$ &-&-\\
&\cellcolor{highlightgreen} DHOG ($\alpha=0$, ablation) &\cellcolor{highlightgreen} $57.49 \pm 0.8929$ &\cellcolor{highlightgreen}$0.5022 \pm 0.0054$ &\cellcolor{highlightgreen} $0.4010 \pm 0.0091$\\
&\cellcolor{highlightblue}DHOG, unsup. ($\alpha=0.05$) &\cellcolor{highlightblue} $\mathbf{66.61 \pm 1.699}$  &\cellcolor{highlightblue} $\mathbf{0.5854 \pm 0.0080}$ &\cellcolor{highlightblue} $\mathbf{0.4916 \pm 0.0160}$\\
&\cellcolor{highlightblue}DHOG, best ($\alpha=0.05$) &\cellcolor{highlightblue} ${66.61 \pm 1.699}$  &\cellcolor{highlightblue} ${0.5854 \pm 0.0080}$ &\cellcolor{highlightblue} ${0.4916 \pm 0.0160}$\\\hline
{{\begin{tabular}{c}\parbox[t]{2mm}{\multirow{8}{*}{\rotatebox[origin=c]{90}{\tiny\textbf{CIFAR-100-20}}}}\end{tabular}}}
& $K$-means on pixels & $13.78 \pm 0.2557$ & $0.0862 \pm 0.0012$ & $0.0274 \pm 0.0005$ \\
& DAC~\cite{chang2017deep} $^\dagger$ & $23.8$ & -& - \\
& DeepCluster~\cite{caron2018deep}  $^\dagger*$ & $18.9$ &-&-\\
& ADC~\cite{haeusser2018associative} & $16.0$ &-&-\\
& IIC~\cite{ji2019invariant} $^\dagger$ & $25.5 \pm 0.462$ &-&-\\
&\cellcolor{highlightgreen} DHOG ($\alpha=0$, ablation) &\cellcolor{highlightgreen} $20.22 \pm 0.2584$ &\cellcolor{highlightgreen} $0.1880 \pm 0.0019$ &\cellcolor{highlightgreen} $0.0846 \pm 0.0026$\\
&\cellcolor{highlightblue} DHOG, unsup. ($\alpha=0.05$) &\cellcolor{highlightblue} $\mathbf{26.05 \pm 0.3519}$  &\cellcolor{highlightblue} $\mathbf{0.2579 \pm 0.0086}$ &\cellcolor{highlightblue} $\mathbf{0.1177 \pm 0.0063}$\\
&\cellcolor{highlightblue}DHOG, best ($\alpha=0.05$) &\cellcolor{highlightblue} ${27.57 \pm 1.069}$  &\cellcolor{highlightblue} ${0.2687 \pm 0.0061}$ &\cellcolor{highlightblue} ${0.1224 \pm 0.0091}$\\ \hline
{{\begin{tabular}{c}\parbox[t]{2mm}{\multirow{4}{*}{\rotatebox[origin=c]{90}{\tiny\textbf{CINIC-10}}}}\end{tabular}}}
&$K$-means on pixels &$20.80 \pm 0.8550$ & $0.0378 \pm 0.0001$ & $0.0205 \pm 0.0007$ \\
&\cellcolor{highlightgreen} DHOG ($\alpha=0$, ablation) &\cellcolor{highlightgreen} $\mathbf{41.66 \pm 0.8273}$ &\cellcolor{highlightgreen} $0.3276 \pm 0.0084$ &\cellcolor{highlightgreen} $\mathbf{0.2108 \pm 0.0034}$\\
&\cellcolor{highlightblue}DHOG, unsup. ($\alpha=0.05$) & \cellcolor{highlightblue}${37.65 \pm 2.7373}$  &\cellcolor{highlightblue} $\mathbf{0.3317 \pm 0.0096}$ &\cellcolor{highlightblue} ${0.1993 \pm 0.0030}$\\ 
&\cellcolor{highlightblue}DHOG, best ($\alpha=0.05$) &\cellcolor{highlightblue} ${43.06 \pm 2.1105}$  &\cellcolor{highlightblue} ${0.3725 \pm 0.0075}$ &\cellcolor{highlightblue} ${0.2396 \pm 0.0087}$\\ \hline
{{\begin{tabular}{c}\parbox[t]{2mm}{\multirow{5}{*}{\rotatebox[origin=c]{90}{\tiny\textbf{SVHN}}}}\end{tabular}}}
&$K$-means on pixels & $11.35 \pm 0.2347$ & $0.0054 \pm 0.0004$ & $0.0007 \pm 0.0004$ \\
&ADC~\cite{haeusser2018associative} & $38.6 \pm 4.1$ &-&-\\
&\cellcolor{highlightgreen} DHOG ($\alpha=0$, ablation) &\cellcolor{highlightgreen} $14.27 \pm 2.8784$ &\cellcolor{highlightgreen} $0.0298 \pm 0.0321$ &\cellcolor{highlightgreen} $0.0209 \pm 0.0237$\\
&\cellcolor{highlightblue}DHOG, unsup. ($\alpha=0.05$) &\cellcolor{highlightblue} $\mathbf{45.81 \pm 8.5427}$  &\cellcolor{highlightblue} $\mathbf{0.4859 \pm 0.1229}$ &\cellcolor{highlightblue} $\mathbf{0.3686 \pm 0.1296}$\\ 
&\cellcolor{highlightblue}DHOG, best ($\alpha=0.05$) &\cellcolor{highlightblue} ${49.05 \pm 8.2717}$  &\cellcolor{highlightblue} ${0.4658 \pm 0.0556}$ &\cellcolor{highlightblue} ${0.3848 \pm 0.0557}$\\ \hline
{{\begin{tabular}{c}\parbox[t]{2mm}{\multirow{11}{*}{\rotatebox[origin=c]{90}{\tiny\textbf{STL-10}}}}\end{tabular}}}
&$K$-means on pixels & $21.58 \pm 0.2151$ & $0.0936 \pm 0.0005$ & $0.0487 \pm 0.0009$ \\
& JULE~\cite{yang2016joint} $^\dagger$ & $27.7$ & - & - \\
& DEC~\cite{xie2016unsupervised} & $35.90$ &-&-\\
& DAC~\cite{chang2017deep} & $46.99$ & $0.3656$ & $0.2565$ \\
& DeepCluster~\cite{caron2018deep}  $^\dagger$ & $33.40$ &-&-\\
&ADC~\cite{haeusser2018associative} & $47.8 \pm 2.7$ &-&-\\
& DCCM~\cite{wu2019deep} & $48.2$ & $0.376$  & $0.262$ \\
&IIC~\cite{ji2019invariant} $^\dagger$ & $\mathbf{59.80 \pm 0.844}$ &-&-\\
& \cellcolor{highlightgreen}DHOG ($\alpha=0$, ablation) &\cellcolor{highlightgreen} $38.70 \pm 4.4696$ &\cellcolor{highlightgreen} $0.3878 \pm 0.0331$ &\cellcolor{highlightgreen} $0.2412 \pm 0.0265$\\
&\cellcolor{highlightblue}DHOG, unsup. ($\alpha=0.05$) &\cellcolor{highlightblue} $48.27 \pm 1.915$  &\cellcolor{highlightblue} $\mathbf{0.4127 \pm 0.0171}$ &\cellcolor{highlightblue} $\mathbf{0.2723 \pm 0.0119}$ \\
&\cellcolor{highlightblue}DHOG, best ($\alpha=0.05$) &\cellcolor{highlightblue} $48.27 \pm 1.915$  &\cellcolor{highlightblue} $0.4127 \pm 0.0171$ &\cellcolor{highlightblue} $0.2723 \pm 0.0119$
\\
\cline{1-5}
\end{tabular}
\vspace{1mm}
\caption{Test set results on all datasets, taken from papers where possible. Results with $^\dagger$ are from \cite{ji2019invariant}. $\mathrm{NMI}(\zRV, \yRV)$ is between remapped predicted label assignments and class targets, $\yRV$. Ablation: DHOG with $\alpha=0.0$ \textbf{is the IIC method but with conditions/augmentations matching those of DHOG($\alpha=0.5$)}. Our method is DHOG with $\alpha=0.05$, highlighted in blue. We give results for the head chosen for the best $\mathrm{NMI}(\zRV, \zRV^\prime)$ and the head chosen for the best $\mathrm{NMI}(\zRV, \yRV)$. In most cases max MI chooses the optimal head.}\label{tab:results}
\end{table*}

Table \ref{tab:results} gives the accuracy, normalised mutual information (NMI), and the adjusted rand index (ARI) between remapped assignments and classification targets. Before assessment a labelling-to-class remapping was computed using the training data and the Hungarian method~\cite{kuhn1955hungarian}. The results listed for DHOG correspond to average over 3 seeded runs. In terms of all measured metrics DHOG outperformed other relevant fully-unsupervised clustering methods, with an accuracy improvement of $4.3\%$ on CIFAR-10, $1.5\%$ on CIFAR-100-20, and $7.2\%$ on SVHN. No Sobel edge-detection was used to account for trivial solutions. The DHOG network converged in half the training time (compared to IIC).

We used a fully unsupervised posthoc head selection according to $\mathrm{NMI}(\zRV, \zRV^\prime)$. The selected heads almost always corresponded with the head that maxmimised $\mathrm{NMI}(\zRV, \yRV)$, where $\yRV$ are user-defined class labels. This means that the DHOG framework produces data groupings that:
\begin{enumerate}
    \item \textbf{Better maximise the widely used MI objective} (between mappings of data augmentations);
    \item Also corresponds better with the challenging \textbf{underlying object classification test objective}.
\end{enumerate}

It is only on STL-10 that DHOG never beat the current state-of-the-art. This may be owing to the need for a STL-10 specific hyper-parameter search. Our aim was to show that the simple hierarchical ordering of heads in DHOG improves performance. The difference between STL-10 with and without the MI cross-head minimisation term (controlled by $\alpha$) shows a marked improvement. Again, note that DHOG uses no preprocessing such as Sobel edge detection to deal with easy solutions to the MI objective.

\begin{figure}[htb]
\begin{center}
\begin{subfigure}{0.35\textwidth}
\includegraphics[width=1\textwidth]{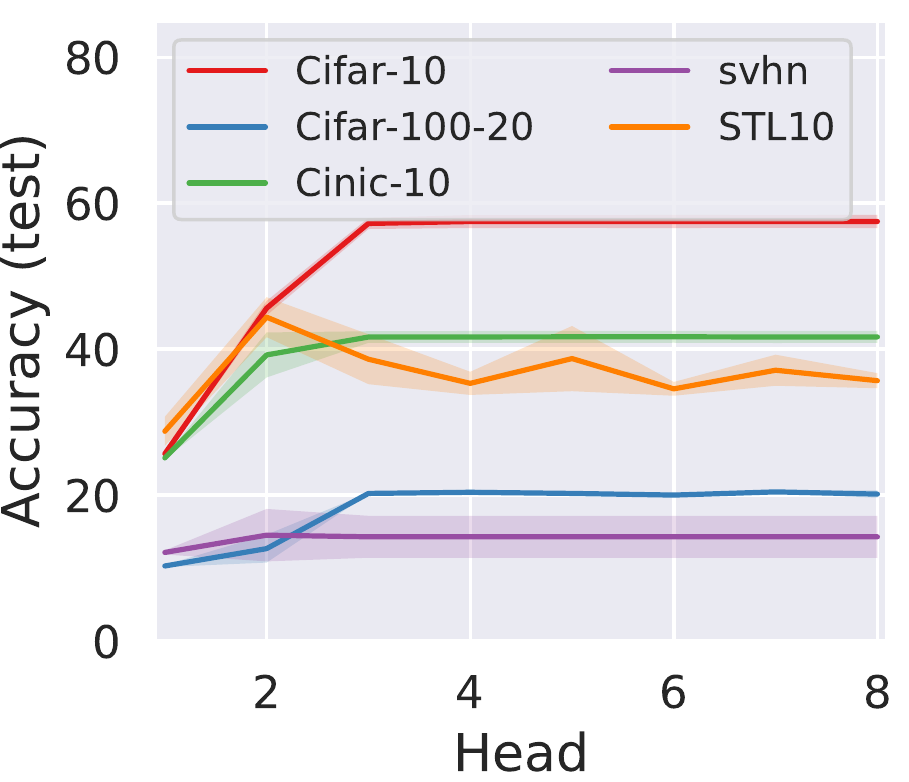}
\caption{Without DHOG training}
\end{subfigure}
\hspace{1cm}
\begin{subfigure}{0.35\textwidth}
\includegraphics[width=1\textwidth]{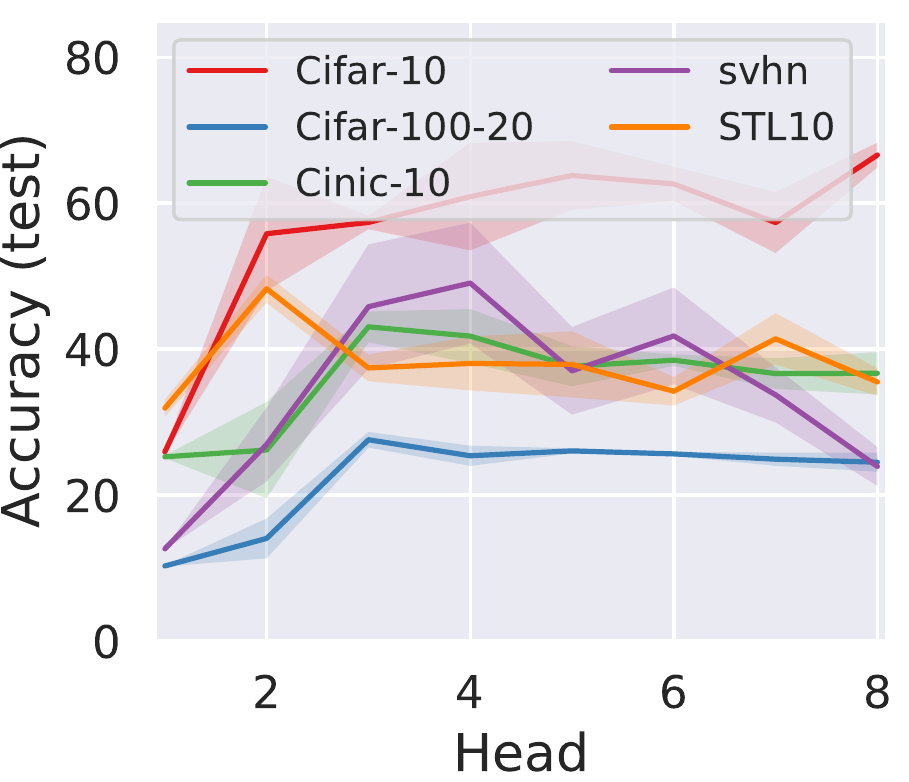}
\caption{With DHOG training}
\end{subfigure}
\end{center}
  \caption{Accuracy per head. DHOG causes heads to learn distinct solutions.}
  
\label{fig:accphead}
\end{figure}

\begin{figure}[htb]
\begin{center}
{\includegraphics[trim={ 0 6mm 0 0},clip,width=0.7\linewidth]{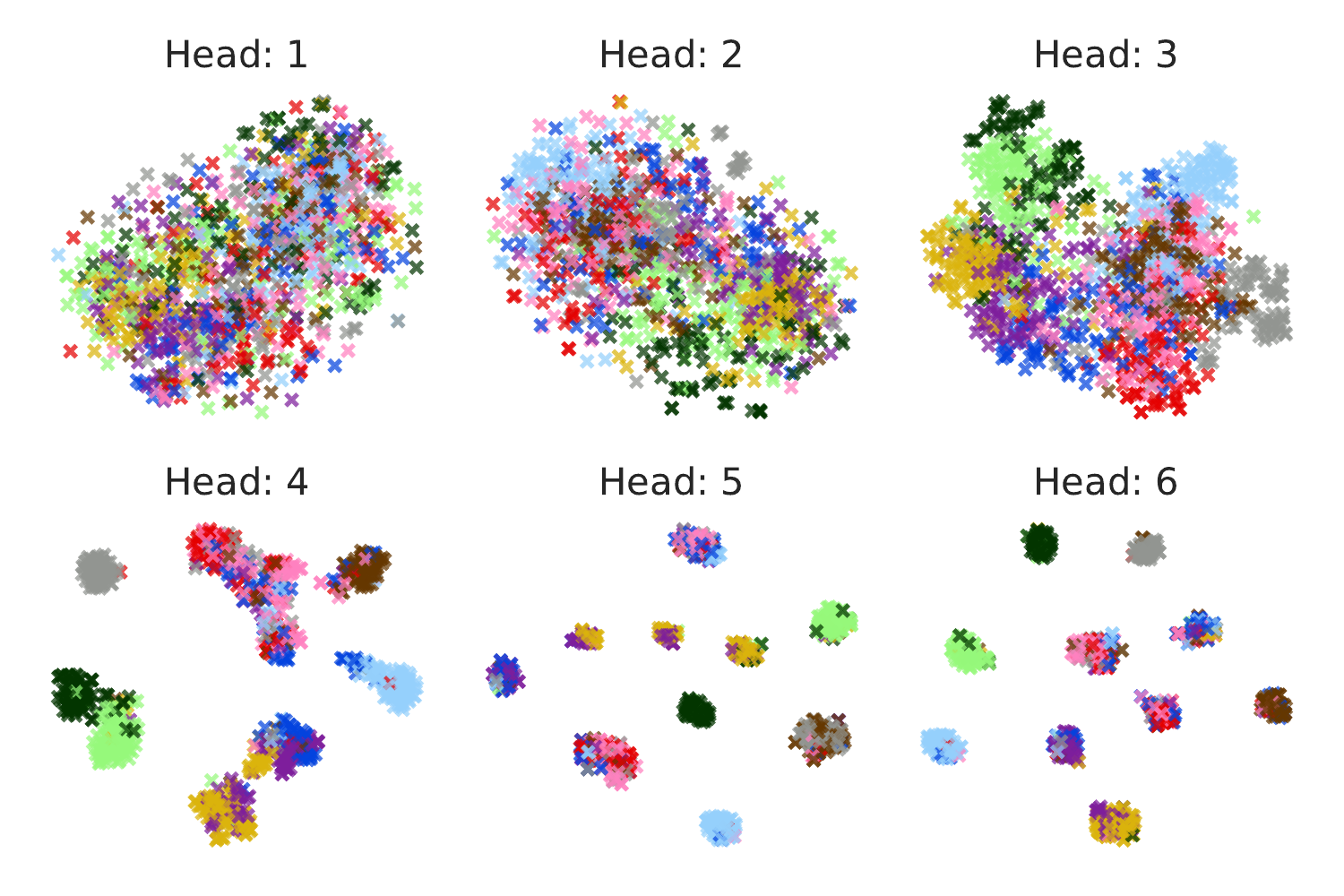}}
\vspace{-5mm}

\end{center}
  \caption{Visualising learned representations via t-SNE plots~\cite{maaten2008visualizing}. Hierarchically ordered representations produce different representations, the clusters of which become more distinct up the hierarchy. \vspace{-6mm}}
  
\label{fig:tsne}
\end{figure}

The advantage of a hierarchical ordering is particularly evident when considering the ablation study: with ($\alpha=0.05$) and without ($\alpha=0$) cross-head MI minimisation. Figure \ref{fig:accphead} (a) and (b) are accuracy versus head curves, showing that without cross-head MI minimisation later heads converge to similar solutions. 

\begin{figure}[!htb]
\begin{center}
\begin{subfigure}{0.49\textwidth}
\includegraphics[width=\linewidth]{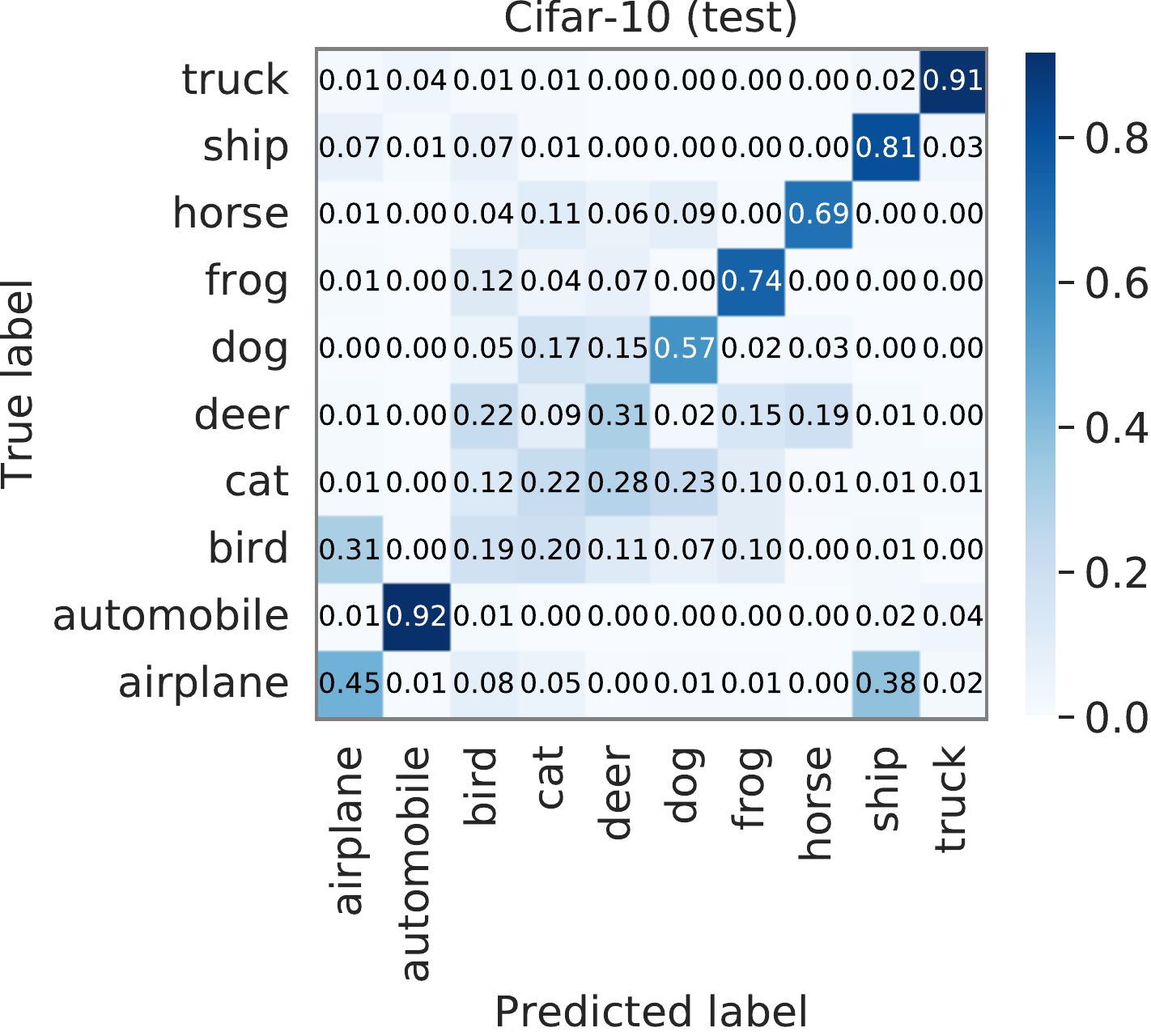}
\caption{Without DHOG training}
\end{subfigure}
\begin{subfigure}{0.49\textwidth}
\includegraphics[width=\linewidth]{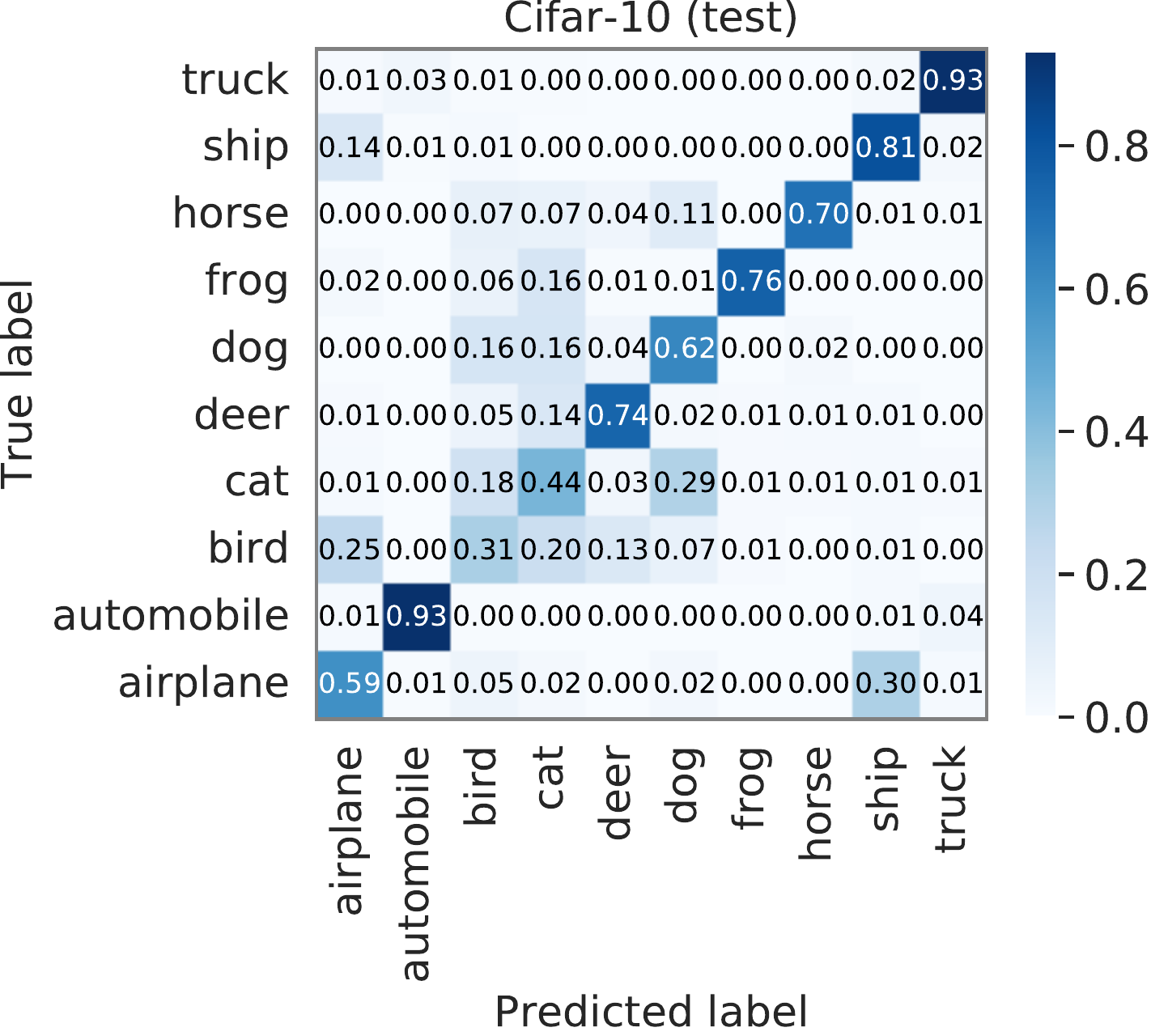}
\caption{With DHOG training}
\end{subfigure}
\end{center}
  \caption{Confusion matrices from the same seed (a) without and (b) with DHOG cross-head MI minimisation. These networks struggle with distinguishing natural objects (birds, cats, deer, etc.), although DHOG does improve this.
\label{fig:confusion} \vspace{-3mm}}

\end{figure}

Figure \ref{fig:tsne} and the confusion matrix in Figure \ref{fig:confusion} (b) show the classes the final learned network confuses in CIFAR-10. Compare this to the confusion matrix in Figure \ref{fig:confusion} (a) where $\alpha=0$ and note the greater prevalence of cross-class confusion. 

Table \ref{tab:images} shows images that yielded the highest probability for each class, and an average of the top 10 images, for both an early ($i=2$) and a late ($i=8$) head. Since the fast auto-augment strategy is broad, the difference between easy-to-compute and complex features is often nuanced. In this case, the grouped images from the earlier head are more consistent in terms of colour or simple patterns. It easier to ascribe notions of `mostly blue background' (class 3) or `light blob with three dark spots' (class 6) for the earlier head. This is also exemplified by the average images: the consistency of images in the early head makes the detail of the averages images clearer, whereas those from the later head are difficult to discern owing to diversity amongst the samples.

\begin{table*}[!htbp]
\def\arraystretch{0.9}
\setlength{\tabcolsep}{0pt}
\centering
\begin{tabular}{|cccccccccc|}

\hline
\multicolumn{10}{|c|}{Predicted labels:}\\
1 & 2 & 3 & 4 & 5 & 6 & 7 & 8 & 9 & 10\\
\hline
\multicolumn{10}{|c|}{Earlier head}\\
\multicolumn{10}{|c|}{\includegraphics[width=1\linewidth, clip, trim={1mm 0 1mm 0}]{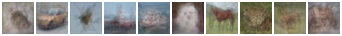}}\\
\multicolumn{1}{|c|}{\includegraphics[width=0.099\linewidth]{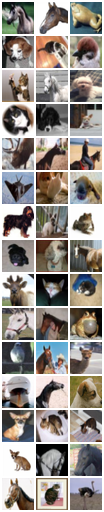}} & \multicolumn{1}{|c|}{\includegraphics[width=0.099\linewidth]{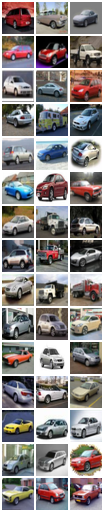}} & \multicolumn{1}{|c|}{\includegraphics[width=0.099\linewidth]{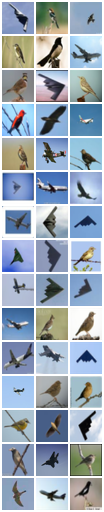}} & \multicolumn{1}{|c|}{\includegraphics[width=0.099\linewidth]{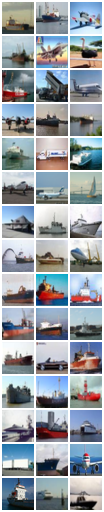}} & \multicolumn{1}{|c|}{\includegraphics[width=0.099\linewidth]{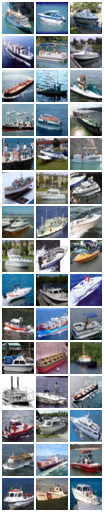}} & \multicolumn{1}{|c|}{\includegraphics[width=0.099\linewidth]{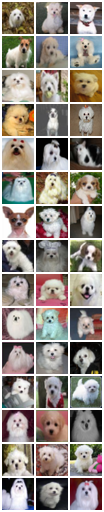}} & \multicolumn{1}{|c|}{\includegraphics[width=0.099\linewidth]{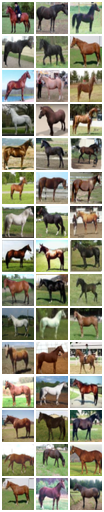}} & \multicolumn{1}{|c|}{\includegraphics[width=0.099\linewidth]{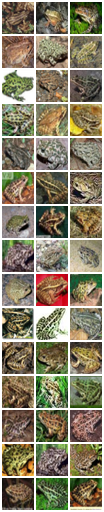}} & \multicolumn{1}{|c|}{\includegraphics[width=0.099\linewidth]{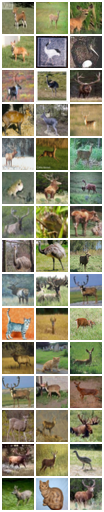}} & \multicolumn{1}{|c|}{\includegraphics[width=0.099\linewidth]{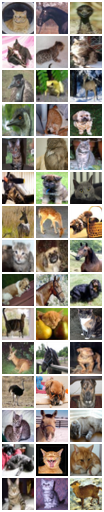} }\\
\hline
\multicolumn{10}{|c|}{Later head} \\
\multicolumn{10}{|c|}{\includegraphics[width=1\linewidth, clip, trim={1mm 0 1mm 0}]{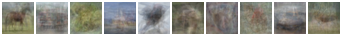}}\\
\multicolumn{1}{|c|}{\includegraphics[width=0.099\linewidth]{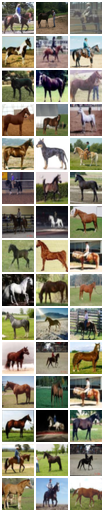}} & \multicolumn{1}{|c|}{\includegraphics[width=0.099\linewidth]{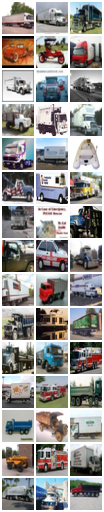}} & \multicolumn{1}{|c|}{\includegraphics[width=0.099\linewidth]{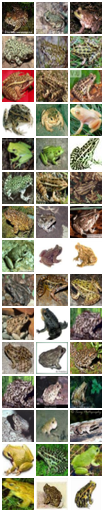}} & \multicolumn{1}{|c|}{\includegraphics[width=0.099\linewidth]{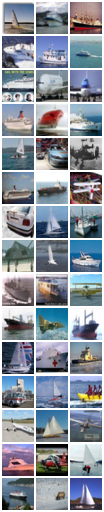}} & \multicolumn{1}{|c|}{\includegraphics[width=0.099\linewidth]{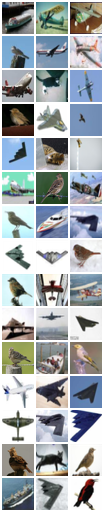}} & \multicolumn{1}{|c|}{\includegraphics[width=0.099\linewidth]{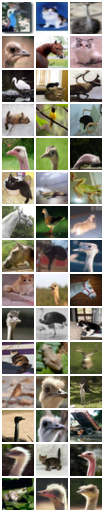}} & \multicolumn{1}{|c|}{\includegraphics[width=0.099\linewidth]{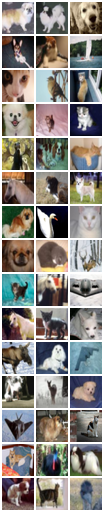}} & \multicolumn{1}{|c|}{\includegraphics[width=0.099\linewidth]{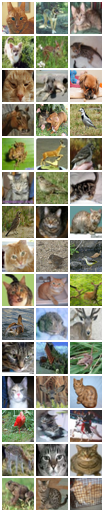}} & \multicolumn{1}{|c|}{\includegraphics[width=0.099\linewidth]{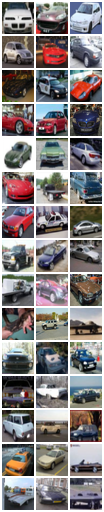}} & \multicolumn{1}{|c|}{\includegraphics[width=0.099\linewidth]{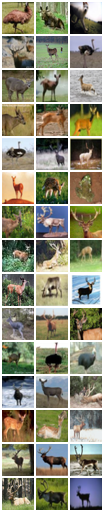} }\\

\hline

\end{tabular}
\vspace{1mm}
\caption{Images that yielded the top probability for each discrete label for an early, $i=2$, and late, $i=8$, head, taken from a single DHOG run on Cifar-10. The average image for the top 10 is also shown. Note particularly the those images grouped by the early head are less diverse than those grouped by the later head. The label associated largely with frogs (8 for the earlier head and 3 for the later head) exemplifies this well.}\label{tab:images}
\end{table*}


\section{Conclusion}\label{sec:conclusion}
We presented deep hierarchical object grouping (DHOG): a method that leverages the challenges faced by current data augmentation-driven unsupervised representation learning methods that maximise mutual information. Learning a good representation of an image using data augmentations is limited by the user, who chooses the set of plausible data augmentations but who is also unable to cost-effectively define an ideal set of augmentations. We argue and show that learning using greedy optimisation typically causes models to get stuck in local optima, since the data augmentations fail to fully describe the sought after invariances to all task-irrelevant information.

We address this pitfall via a simple-to-complex ordered sequence of representations. DHOG works by minimising mutual information between these representations such that those later in the hierarchy are encouraged to produce unique and independent discrete labellings of the data (w.r.t.~earlier representations). Therefore, later heads avoid becoming stuck in the same local optima of the original mutual information objective (between augmentations, applied separately to each head). Our tests showed that DHOG resulted in an improvement of $4.2\%$ on CIFAR-10, $1.5\%$ on CIFAR-100-20, and $7.2\%$ on SVHN, without using preprocessing such as Sobel edge detection.

\section*{Acknowledgements}

This work was supported in part by the EPSRC Centre for Doctoral Training in Data Science, funded by the UK Engineering and Physical Sciences Research Council (grant EP/L016427/1) and the University of Edinburgh.

This research was part funded from a Huaweil DDMPLab Innovation Research Grant DDMPLab5800191.

\bibliographystyle{splncs04}
\bibliography{eccv202submission}
\end{document}